\documentclass[letterpaper, 10 pt, conference]{ieeeconf}  % Comment this line out if you need a4paper

\IEEEoverridecommandlockouts                              % This command is only needed if 
\usepackage{color}
\usepackage[font=small]{caption}
\usepackage{siunitx}
\usepackage{mathtools}
\usepackage{amssymb} 
\usepackage{gensymb}

\usepackage{subcaption, graphicx}  % for png and subfigure
\usepackage[utf8]{inputenc}
\usepackage{pgfplots}
\usepackage{tikz}
\usepackage{tikzscale}
\usetikzlibrary{calc,trees,positioning,fit,arrows,decorations.pathreplacing}

\usepackage[breaklinks=true,bookmarks=false]{hyperref}

\newcommand{\etal}{\textit{et al}.~}
\newcommand{\ie}{\textit{i}.\textit{e}.}

\newcommand\figref[1]{Fig.~{\ref{#1}}}
\newcommand\tabref[1]{Table~{\ref{#1}}}
\newcommand{\norm}[1]{\left\lVert #1 \right\rVert_2}

\DeclareMathOperator*{\argmax}{arg\,max}

\title{\LARGE \bf
3D Human Pose Estimation in RGBD Images for Robotic Task Learning
}

\author{Christian Zimmermann*, Tim Welschehold*, Christian Dornhege, Wolfram Burgard and Thomas Brox% <-this % stops a space
\thanks{*Indicates equal contribution. All authors are with the Department of Computer Science at the University
of Freiburg, 79110 Freiburg, Germany. This work was supported by the Baden-Wurttemberg Stiftung as part of the projects ROTAH and RatTrack.}% <-this % stops a space
}

\begin{document}

\maketitle
\thispagestyle{empty}
\pagestyle{empty}

%%%%%%%%%%%%%%%%%%%%%%%%%%%%%%%%%%%%%%%%%%%%%%%%%%%%%%%%%%%%%%%%%%%%%%%%%%%%%%%%
\begin{abstract}
We propose an approach to estimate 3D human pose in real world units from a single RGBD image and show that it exceeds performance of monocular 3D pose estimation approaches from color as well as pose estimation exclusively from depth. Our approach builds on robust human keypoint detectors for color images and incorporates depth for lifting into 3D. We combine the system with our learning from demonstration framework to instruct a service robot without the need of markers. Experiments in real world settings demonstrate that our approach enables a PR2 robot to imitate manipulation actions observed from a human teacher.
\end{abstract}

%%%%%%%%%%%%%%%%%%%%%%%%%%%%%%%%%%% MAIN PART %%%%%%%%%%%%%%%%%%%%%%%%%%%%%%%%%%%%%%%%%%%%%
\section{INTRODUCTION}
% SECTION: Introduction
Perception and understanding of the surrounding environment is vital for many robotics tasks. Tasks involving interaction with humans heavily rely on prediction of the human location and its articulation in space. These applications involve, e.g., gesture control, hand-over maneuvers, and learning from demonstration.

On the quest of bringing service robots to mass market and into common households, one of the major milestones is their instructability: consumers should be able to teach their personal robots their own custom tasks. Teaching should be intuitive and not require expert knowledge or programming skills. Ideally, the robot should learn from observing its human teacher demonstrating the task at hand. Hence it needs to be able to follow the human motion. Especially the hands play a key role as they are our main tool of interaction with the environment. 

Estimation of human pose is challenging due to variation in appearance, strong articulation and heavy occlusions by themselves or objects. Recent approaches present robust pose estimators in 2D, but for robotic applications full 3D pose estimation in real world units is indispensable. In this paper, we bridge this gap by lifting 2D predictions into 3D while incorporating information from a depth map. This lifting via a depth map is non-trivial for multiple reasons, for instance, occlusion of the person by an object leads to misleading depths, see Fig.~\ref{fig:qualitative_results}.

We present a learning based approach that predicts full 3D human pose and hand normals from RGBD input. It outperforms existing baseline methods and we show feasibility of teaching a robot tasks by demonstration.

The approach first predicts human pose in 2D given the color image. A deep network takes the 2D pose and the depth map as input and derives the full 3D pose from this information. Building on the predicted hand locations we additionally infer the hand palm normals from the cropped color image. Based on this pose estimation system, we demonstrate the feasibility of our action learning from human demonstration approach without the use of artificial markers on the person. We reproduce the demonstrated actions  on our robot in real world experiments. An implementation of our approach and a summarizing video are available online.\footnote{\href{https://lmb.informatik.uni-freiburg.de/projects/rgbd-pose3d/}{https://lmb.informatik.uni-freiburg.de/projects/rgbd-pose3d/}}

\begin{figure}
\centering
\begin{subfigure}{.4\columnwidth}
  \centering
  \includegraphics[width=\linewidth]{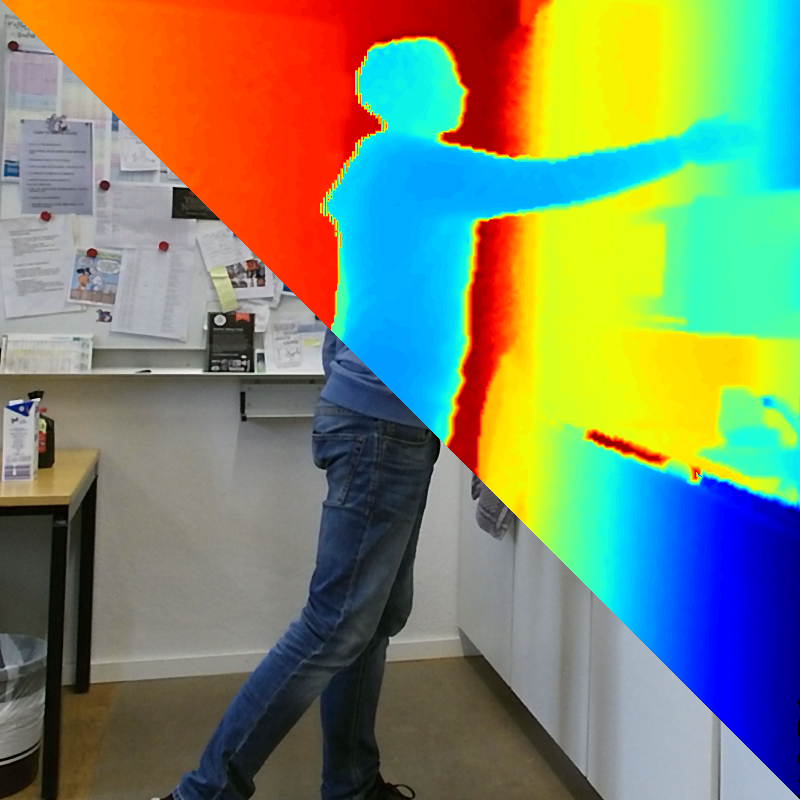}
\end{subfigure}
\begin{subfigure}{.55\columnwidth}
  \centering
  \includegraphics[width=\linewidth]{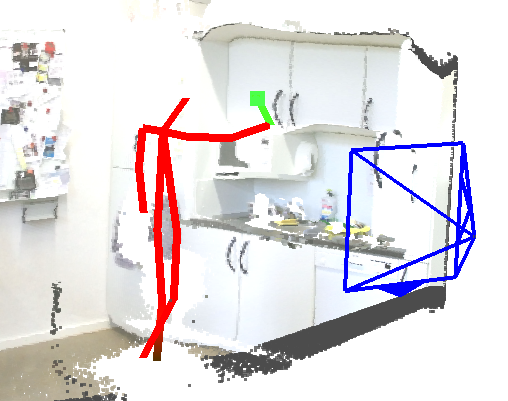}
\end{subfigure}
\caption{Given a color image and depth map, our system detects keypoints in 3D and predicts the normal vectors of the hands if visible. Predictions of that system enable us to teach a robot tasks by demonstration.}
\label{fig:teaser}
\end{figure}

\section{RELATED WORK}
\begin{figure*}
\centering
    \includegraphics[]{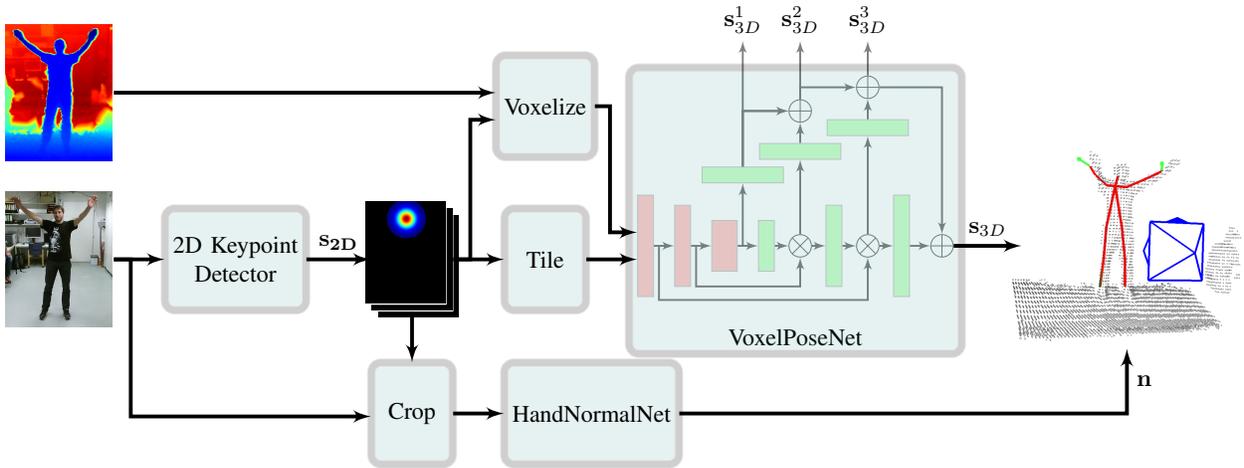}
\caption{First, we predict the keypoint locations in the color image. The predicted score maps are tiled along the z-dimension and a person centered occupancy voxel grid is calculated from the depth map. Based on these inputs \textit{VoxelPoseNet} predicts keypoints in 3D. Cropped images around the hand are fed to \textit{HandNormalNet}, which predicts the normals. Red and green blocks represent convolutional and deconvolutional operations. Concatenation is denoted by $\otimes$ and $\oplus$ is the elementwise add operation.}\label{fig:overview_pose}
\end{figure*}

% SECTION: Related work
The vast majority of publications in the field of human pose estimation deal with the problem of inferring keypoints in 2D given a color image \cite{cao2017realtime, wei2016cpm}, which is linked to the availability of large scale datasets \cite{andriluka_2d_2014, lin2014microsoft}. Due to the large datasets, networks for keypoint localization in 2D have reached impressive performance, which we integrate into our approach.

Recent techniques learn a prior for human pose that allows prediction of the most likely 3D pose given a single color image \cite{martinez_simple_2017, tome_lifting_2017}. Predictions of most monocular approaches live in a scale and translation normalized frame, which makes them impracticable for many robotic applications. Approaches that can recover full 3D from RGB alone \cite{mehta_vnect_2017} use assumptions to resolve the depth ambiguity. Our approach does not need any assumptions to predict poses in world coordinates.

All approaches that provide predictions in real world units are based on active depth sensing equipment. Most prominent is the Microsoft Kinect v1 sensor. Shotton~\etal\cite{shotton2013real} describes a discriminative method that is based on random forest classifiers and yields a body part segmentation. This work was followed by numerous approaches that propose using random tree walks \cite{yub2015random}, a viewpoint invariant representation \cite{haque_towards_2016} or local volumetric convolutional networks for local predictions \cite{moon_holistic_2017}. In contrast to the mentioned techniques, we incorporate depth and color in a joint approach.
So far little research went into approaches that incorporate both modalities \cite{buys2014adaptable}. We propose a deep learning based approach to combine color and depth. Our approach leverages the discriminative power of keypoint detectors trained on large scale databases for color images and complements them with information from the depth map for lifting to real world 3D coordinates.

In the field of learning from demonstration, Calinon~\etal\cite{Calinon12Hum} use markers to track human hand trajectories for action learning. M\"{u}hlig~\etal\cite{raey} use an articulated model of the human body to track teacher actions. Although being able to imitate the human manipulation motions, grasp poses on the objects are either pre-programmed or assumed as given. Mao~\etal\cite{mao2014learning} use a marker-less hand tracking method to teach manipulation tasks. Unlike our work, they assume that the human demonstrations are suitable for robot execution without further adjustment.

\section{APPROACH}
% SECTION: APPROACH

In this work we aim to estimate 3D human poses and the hand normal vectors from RGBD input. This procedure is summarized in \figref{fig:overview_pose}. 
Subsequently, we extract human motion trajectories from demonstrations and transfer them to the robot with regard to its kinematics and grasping capabilities.

\subsection{Human Pose Estimation}
\label{subsec:human_pose_estimation}

% formal problem definition
We aim for estimating the human body keypoints $\mathbf{w} = (\vec{w}_1, \dots, \vec{w}_J) \in \mathbb{R}^{3 \times J}$ for $J$ keypoints in real world coordinates relative to the Kinect sensor given color image $\mathbf{I} \in \mathbb{R}^{N\times M\times 3}$, depth map $\mathbf{D}' \in \mathbb{R}^{N'\times M'}$ and their calibration. Additionally we predict the hand normal vectors $\mathbf{n} \in \mathbb{R}^{3\times 2}$ for both hands of the person. Without loss of generality we define the coordinate system, our predictions live in, to be identical with the color sensors frame. 

% warped depth map
For the Kinect, the color and depth sensors are located in close proximity, but still the frames resemble two distinct cameras. Our approach needs to collocate information of the two frames. Therefore we transform the depth map into the color frame using the camera calibration. As a result, our approach operates on the warped depth map $\mathbf{D} \in \mathbb{R}^{N\times M}$. Due to occlusions, differences in resolution and noise, the resulting depth map $\mathbf{D}$ is sparse, but for better visualization a linear interpolation of $\mathbf{D}$ is shown in \figref{fig:overview_pose}.

\subsubsection{Color Keypoint Detector}
% ColorPose2D: Task definition
The keypoint detector is applied to the color image $\mathbf{I}$, which yields score maps $\mathbf{s_{\text{2D}}} \in \mathbb{R}^{N\times M\times J}$ encoding the likelihood of a specific human keypoint being present. 
The maxima of the score maps $\mathbf{s}_{\text{2D}}$ correspond to the predicted keypoint locations $\mathbf{p} = (\vec{p}_0, \dots \vec{p}_J) \in \mathbb{R}^{2\times J}$ in the image plane.
% ColorPose2D: What do we use
Thanks to many datasets with annotated color frames for human pose estimation \cite{lin2014microsoft, andriluka_2d_2014}, robust detectors are available. We use the Open Pose Library \cite{cao2017realtime, simon2017hand, wei2016cpm} with fixed weights in this work.

\subsubsection{VoxelPoseNet}
% Voxelization
Given the warped depth map $\mathbf{D}$ a voxel occupancy grid $\mathbf{V} \in \mathbb{R}^{K\times K\times K}$ is calculated with $K=64$. For this purpose the depth map $\mathbf{D}$ is transformed into a point cloud and we calculate an 3D coordinate $\vec{w}_{\text{r}}$, which is the center of $\mathbf{V}$. We calculate $\vec{w}_{\text{r}}$ as back projection of the predicted 2D 'neck' keypoint $\vec{p}_{\text{r}}$ using the median depth $d_r$ extracted from the neighborhood of $\vec{p}_{\text{r}}$ in $\mathbf{D}$: 

\begin{equation}
    \vec{w}_r = d_r \cdot \mathbf{K}^{-1} \cdot \vec{p}_r \text{.}
\end{equation}
Where $\mathbf{K}$ denotes the intrinsic calibration matrix camera and $\vec{p}_r$ is in homogeneous coordinates. We pick the value $d_r$ from the depth map taking into account the closest $3$ neighboring valid depth values around $\vec{p}_r$.
We calculate $\mathbf{V}$ by setting elements to $1$, when there is at least one point of the point cloud lying in the interval represented and zero otherwise. We chose the resolution of the voxel grid to be approximately \SI{3}{cm}.

% VoxelPoseNet: Whats the representation?
\textit{VoxelPoseNet} gets $\mathbf{V}$ and a volume of tiled score maps $\mathbf{s}_{\text{2D}}$ as input and processes them with a series of 3D convolutions. We propose to tile $\mathbf{s}_{\text{2D}}$ along the z-axis, which is equivalent to an orthographic projection approximation. 
% VoxelPoseNet: Whats the output?
\textit{VoxelPoseNet} estimates score volumes $\mathbf{s}_{\text{3D}} \in \mathbb{R}^{K\times K\times K\times J}$, which resemble keypoint likelihoods the same way as its 2D counterpart
\begin{equation}
    \mathbf{w}_{\text{VPN}} = \argmax_{x, y, z}(\mathbf{s_{\text{3D}}}) \text{.}
\end{equation}
% Assembling final prediction
We use the following heuristic to assemble our final prediction: On the one hand $\mathbf{w}_{\text{VPN}}$ is predicted by \textit{VoxelPoseNet}. On the other hand we take the z-component of $\mathbf{w}_{\text{VPN}}$ and the predicted 2D keypoints $\mathbf{p}_{\text{2D}}$ to calculate another set of world coordinates $\mathbf{w}_{\text{projected}}$. For these coordinates the accuracy in x- and y-direction is not limited by the choice of $K$ anymore. We chose our final prediction $\mathbf{w}$ from $\mathbf{w}_{\text{projected}}$ and $\mathbf{w}_{\text{VPN}}$ based on the 2D networks prediction confidence, which is the score of $\mathbf{s}_{\text{2D}}$ at $\mathbf{p}$. 

% VoxelPoseNet: What layers does our network have?
\figref{fig:overview_pose} shows the network architecture used for \textit{VoxelPoseNet}, which is a encoder decoder architecture inspired by the U-net \cite{ronneberger2015u} that uses dense blocks \cite{huang2017densely} in the encoder. While decoding to the full resolution score map, we incorporate multiple intermediate losses denoted by $\mathbf{s}^i_\text{3D}$, which are discussed in section section \ref{subsec:network_loss}.

\subsection{Hand Normal Estimation}
%Task
The approach presented in section section \ref{subsec:human_pose_estimation} yields locations for the human hands, which are used to crop the input image centered around the predicted hand keypoint.
% Architecture overview
For \textit{HandNormalNet} we adopt our previous work on hand pose estimation \cite{zb2017hand}. We exploit that the network from \cite{zb2017hand} estimates the relative transformation between the depicted hand pose and a canonical frame, which gives us the normal vector. We use that network without further retraining.

\subsection{Network training}
\label{subsec:network_loss} 
% Loss functions
We train \textit{VoxelPoseNet} using a sum of squared $L_2$ losses:
\begin{equation}
    \text{L} = \sum_{i} \norm{\mathbf{s}^\text{gt}_\text{3D} - \mathbf{s}^{i \text{, pred}}_\text{3D}}^2
\end{equation}
with a batch size of $2$. Datasets used for training are discussed in section section \ref{sec:datasets}. 
The networks are implemented in Tensorflow \cite{abadi2016tensorflow} and we use the ADAM solver \cite{adam_kingsma}. We train for $40000$ iterations with an initial learning rate of $10^{-4}$, which drops by the factor $0.1$ every $10000$ iterations. 
% How i created GT
Ground truth score volumes $\mathbf{s}^\text{gt}_\text{3D}$ are calculated from the ground truth keypoint location within the voxel $\mathbf{V}$. A Gaussian function is placed at the ground truth location and normalized such that its maximum is equal to $1$.

\subsection{Action learning}
With the ability to record the human motion trajectories, action learning requires them to be transferred to the robot. Due to its deviating kinematics and grasping capabilities the robot cannot directly reproduce the human motions. For the necessary adaption and the action model generation we use the learning-from-demonstration approach presented in our previous work~\cite{twelsche16iros,twelsche17iros}. Here, the robot motion is designed to follow the teacher's demonstrations as closely as possible, while deviating as much as necessary to fulfill constraints posed by its geometry. We pose it as a graph optimization problem, in which trajectories of the manipulated object and the teacher's hand and torso serve as input. We account for the robot's grasping skills and kinematics as well as occlusions in the observations and collisions with the environment. We assume that the grasp on the object is fixed during manipulation and all trajectories are smooth in the sense that consecutive poses should be near each other. These constraints are addressed via the graphs edges. During optimization the teacher's demonstrations are adapted towards trajectories that are feasible for robot execution. For details on the graph structure and the implementation we refer to Welschehold~\etal~\cite{twelsche16iros,twelsche17iros}.

\section{DATASETS}
\label{sec:datasets}
 % SECTION: Datasets
 Currently there are no datasets for the Kinect v2 that provide high-quality skeleton annotation of the person. Due to its long presence, most publicly available sets are recorded with the Kinect v1.
 These datasets are not suited for our scenario, because of major technical differences between the two models.
 More recently published datasets, such as Shahroudy~\etal\cite{Shahroudy_2016_CVPR}, transitioned to the new model but used the Kinect SDK's prediction as pseudo ground truth. Using those datasets is prohibitive for exceeding the Kinect SDK's performance.

\def\datasetFigureWidht{0.225\columnwidth}
\begin{figure}
\centering
\begin{subfigure}{\datasetFigureWidht}
  \centering
  \includegraphics[width=\textwidth]{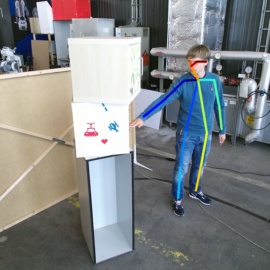}
\end{subfigure}
\begin{subfigure}{\datasetFigureWidht}
  \centering
  \includegraphics[width=\textwidth]{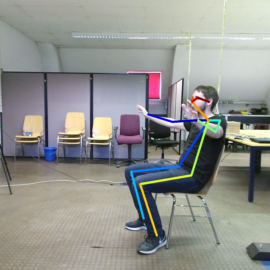}
\end{subfigure}
\begin{subfigure}{\datasetFigureWidht}
  \centering
  \includegraphics[width=\textwidth]{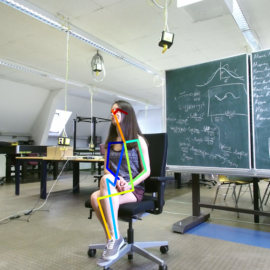}
\end{subfigure}
\begin{subfigure}{\datasetFigureWidht}
  \centering
  \includegraphics[width=\textwidth]{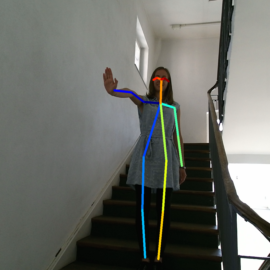}
\end{subfigure}
\caption{Examples from the \textit{MKV} dataset with ground truth skeleton overlayed. The two leftmost ones are samples from the training set and the other two show the evaluation set. }\label{fig:mkv_samples}
\end{figure}

\subsection{Multi View Kinect Dataset (MKV)}
\label{subsec:kinect_dataset}
Therefore, for training of our neural network we recorded a new dataset, which comprises 5 actors, 3 locations, and up to 4 viewpoints. There are 2 female and 3 male actors and the locations resemble different indoor setups. Some examples are depicted in \figref{fig:mkv_samples}. The poses include various upright and sitting poses as well as walking sequences.
Short sequences were recorded simultaneously by multiple calibrated Kinect v2 devices with a frame rate of \SI{10}{Hz}, while recording the skeletal predictions of the Kinect SDK. In a post processing step we applied state-of-the-art Human Keypoint Detectors  \cite{cao2017realtime, simon2017hand, wei2016cpm} and used standard triangulation techniques to lift the 2D predictions into 3D.
This results in a dataset with $22406$ samples. Each sample comprises of color image, depth map, infrared image, the SDK prediction and a ground truth skeleton annotation we get through triangulation. The skeleton annotations comprises of $18$ keypoints that follow the Coco definitions \cite{lin2014microsoft}.
We apply data augmentation techniques and split the set into an evaluation set of $3546$ samples (\textit{MVK-e}) and a training set with $18860$ (\textit{MVK-t}). We divide the two sets by actors and assign both female actors into the evaluation set, which also leaves one location unique to this set. 
Additionally this dataset contains annotated hand normals for a small subset of the samples. The annotations stem from detected and lifted hand keypoints, which were used to calculate the hand normal ground truth. Because detection accuracy was much lower and bad samples were discarded afterwards this dataset is much smaller and provides a total of $129$ annotated samples.

\subsection{Captury Dataset}
Due to the limited number of cameras in the \textit{MKV} setup and the necessity to avoid occluding too many cameras views at the same time, we are limited in the amount of possible object interaction of the actors. Therefore we present a second dataset that was recorded using a commercial marker-less motion capture system called Captury\footnote{http://www.thecaptury.com}. It uses $12$ cameras to track the actor with \SI{120}{Hz} and we calibrated a Kinect v2 device with respect to the Captury. The skeleton tracking provides $23$ keypoints, from which we use $13$ for comparison. We recorded three actors, which performed simple actions like pointing, walking, sitting and interacting with objects like a ball, chair or umbrella. One actor of this setting was already recorded for the \textit{MKV} dataset and therefore constitutes the set used for training. Two previously unseen actors were recorded and form the evaluation set. There are $1535$ samples for training (\textbf{CAP-t}) and $1505$ samples for evaluation (\textbf{CAP-e}).
The definition of human keypoints between the two datasets is compatible, except for the "head" keypoint, which misses a suitable counterpart in the \textit{MKV} dataset. This keypoint is excluded from evaluation to avoid systematic error in the comparison. 

 % SECTION: EXPERIMENTS
\section{EXPERIMENTS - POSE ESTIMATION}

%% subsection
\subsection{Datasets for training}
\tabref{tab:train_sets} shows that the proposed \textit{PoseNet3D} already reaches good results on the evaluation split of both datasets when trained only on \textit{MKV-t}. Training a network only on \textit{CAP-t} leads to inferior performance, which is due to starkly limited variation in the training split of the Captury dataset, which only contains a single actor and scene. Training jointly on both sets performs roughly on par with training exclusively on \textit{MKV-t}. Therefore we use \textit{MKV-t} as default training set for our networks and evaluate on \textit{CAP-e} for following experiments. Furthermore, we confirm generalization of our \textit{MKV-t} trained approach on the \textit{InOutDoor} Dataset \cite{mees2016choosing}. Because the dataset does not contain pose annotations we present qualitative results in the supplemental video.

\begin{table}
\begin{center}
\begin{tabular}{|c|c|c|c|}
\hline
 Training set & \textit{CAP-e} full & \textit{CAP-e} subset & \textit{MKV-e} \\
\hline\hline
\textit{MKV-t}                      & $0.627$  & $0.618$  & $0.793$\\
\textit{CAP-t}                      & $0.603$  & $0.588$  & $0.665$\\
\textit{CAP-t} \& \textit{MKV-t}    & $0.633$  & $0.625$  & $0.794$\\
\hline
\end{tabular}
\caption{Performance measured as area under the curve (AUC) for different training sets of \textit{VoxelPoseNet}. \textit{CAP-t} does not generalize to \textit{MKV-e}, whereas \textit{MKV-t} provides sufficient variation to generalize to \textit{CAP-e}. Training jointly on \textit{CAP-t} and \textit{MKV-t} doesn't improve results much anymore. }\label{tab:train_sets}
\end{center}
\end{table}

%% subsection
\subsection{Comparison to literature}
\begin{figure}
    \centering  
        \includegraphics[width=\columnwidth, height=.6\columnwidth]{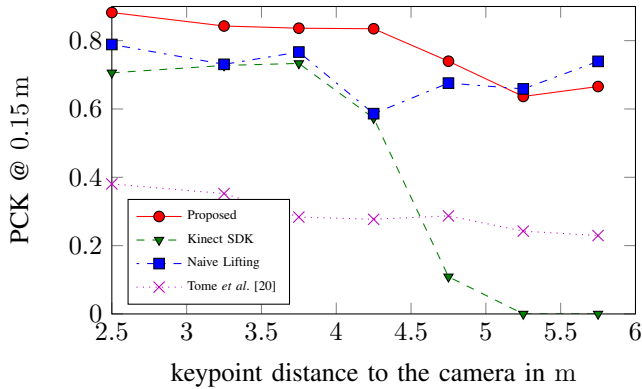}
\caption{Percentage of correct keypoints (PCK) over their distance to the camera. Most approaches are only mildly affected by the keypoint distance to the camera, but the Kinect SDK can only provide predictions in a limited range.}\label{fig:pck_over_dist}
\end{figure}

\begin{figure*}
\centering
    \includegraphics[]{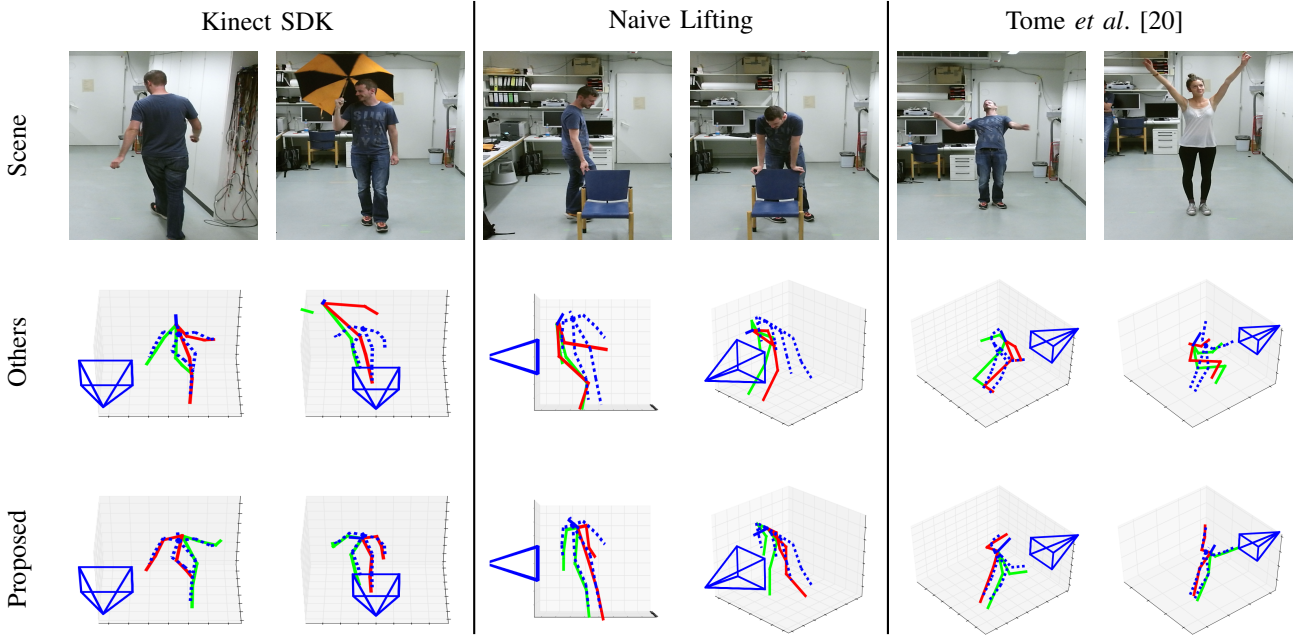}
\caption{Typical failure cases of the algorithms evaluated for samples from \textit{CAP-e}. The first row shows the scene and the other two rows depict the ground truth skeleton in dashed blue and the prediction in solid green and red. Green color indicates the persons right side. Predictions of our proposed approach are shown in the last row, whereas the middle row shows predictions by other algorithms. The first two columns correspond to predictions of the Kinect SDK, the next two are by the Naive Lifting approach and the last two by the approach presented by Tome \etal\cite{tome_lifting_2017}. Typical failures for the SDK are caused by objects and or people that face away from the camera. Naive Lifting fails when any sort of keypoint occlusion is present. }
\label{fig:qualitative_results}
\end{figure*}

In \tabref{tab:epe_results} we compare our approach with common baseline methods. 
The first baseline is the Skeleton Tracker integrated in Microsofts Software Development Kit\footnote{https://www.microsoft.com/en-us/download/details.aspx?id=44561} (Kinect SDK). We show that its performance heavily drops on the more challenging subset and therefore argue that it is unsuitable for many robotics applications. Furthermore, \figref{fig:pck_over_dist} shows that the Kinect SDK is unable to predict keypoints farther away than a certain distance. The qualitative examples in \figref{fig:qualitative_results} reveal that the SDK is led astray by objects and is unable to distinguish if a person is facing towards or away from the camera, which expresses itself in mixing up left and right side.

\begin{table}
\begin{center}
\begin{tabular}{|c|c|c|c|}
\hline
 & Captury full & Captury subset & Multi Kinect \\
\hline\hline
Kinect SDK & $13.5$ & $16.4$ & $8.9$\\
Naive Lifting & $14.7$ & $15.2$ & $8.8$\\
Tome \etal\cite{tome_lifting_2017} & $22.7$ & $21.9$ & $15.1$\\

Proposed & $\mathbf{11.2}$ & $\mathbf{11.6}$ & $\mathbf{6.1}$\\
\hline
\end{tabular}
\caption{Average mean end point error per keypoint of the predicted 3D pose for different approaches in \SI{}{cm}. For the Captury dataset we additionally report results on the subset of non-frontal scenes and with object interaction.}\label{tab:epe_results}
\end{center}
\end{table}

The baseline named Naive Lifting uses the same Keypoint detector for color images as our proposed approach and simply picks the corresponding depth value from the depth map. It chooses the depth value as median value of the $3$ closest neighbors. The approach shows reasonable performance, but is prone to pick bad depth values from the noisy depth map. Also any kind of occlusion results into an error, which is seen in \figref{fig:qualitative_results}.

\begin{figure}
    \centering  
        \includegraphics[width=\columnwidth, height=.6\columnwidth]{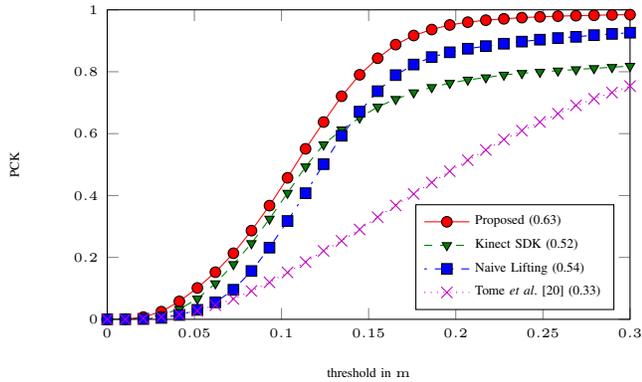}
\caption{Performance of different algorithms on \textit{CAP-e} measured as percentage of correct keypoints (PCK) on the more challenging subset of non-frontal poses and object interaction.}\label{fig:auc_curves_cap}
\end{figure}

Tome \etal\cite{tome_lifting_2017} predicts scale and translation normalized poses. So in order to compare the results to the other approaches we provide the algorithm with ground truth scale and translation. For every prediction we seek scale and translation in order to minimize the reconstruction error between ground truth and prediction. \tabref{tab:epe_results} shows that the approach reaches competitive results, but performs worst in our comparison, which is reasonable given the lack of depth information. In \figref{fig:auc_curves_cap} the approach stays far behind, which partly lies in the fact that the approach misses to provide predictions in $8.7 \%$ of the frames of \textit{CAP-e}, which compares to $12.4 \%$ for Kinect SDK and $~0 \%$ for Naive Lifting and our approach.

\textit{VoxelPoseNet} outperforms its baseline methods, because it exploits both modalities. On the one hand, color information helps to disambiguate left and right side, which is infeasible from depth alone. On the other hand, the depth map provides valuable information to exactly infer the 3D keypoint. Furthermore, the network learns a prior about possible body part configurations, which makes it possible to infer 3D locations even for completely occluded keypoints (see \figref{fig:qualitative_results}).

%% subsection
\subsection{HandNormalNet}
We use the annotated samples of \textit{MKV} to evaluate the accuracy of the normal estimation we achieve with the adopted network from \cite{zb2017hand}. For the $129$ samples we get an average angular error of $60.3$ degree, which is sufficient for the task learning application as is shown in the next section.

\section{EXPERIMENTS - ACTION LEARNING}
\label{exp_action_learning}
We evaluate our recently proposed graph-based approach~\cite{twelsche17iros} for learning a mobile manipulation task from human demonstrations on data acquired with the approach for 3D human pose estimation presented in this work. We evaluate the methods on the same four tasks as in our previous work~\cite{twelsche17iros}: one task of opening and moving through a room door and three tasks of opening small furniture pieces. The tasks will be referred to as room door, swivel door, drawer, and sliding door. Each consists of three parts. First a specific part of the object is grasped, \ie, a handle or a knob, then the object is manipulated according to its geometry, and lastly released. The demonstrations were recorded with a Kinect v2 at $\SI{10}{Hz}$. As we need to track both, the manipulated object and the human teacher, the actions were recorded from a perspective that show the human from the side or back making pose estimation challenging. For an example of the setup see \figref{fig:robotaction}.   

\subsection{Adapting Human Demonstrations to Robot Requirements}
\begin{figure}[t]
	\centering
	\begin{tabular}{cc}  		
  	\includegraphics[width=0.239\textwidth ,trim={10.0cm 2.5cm 5.5cm 3.5cm},clip]{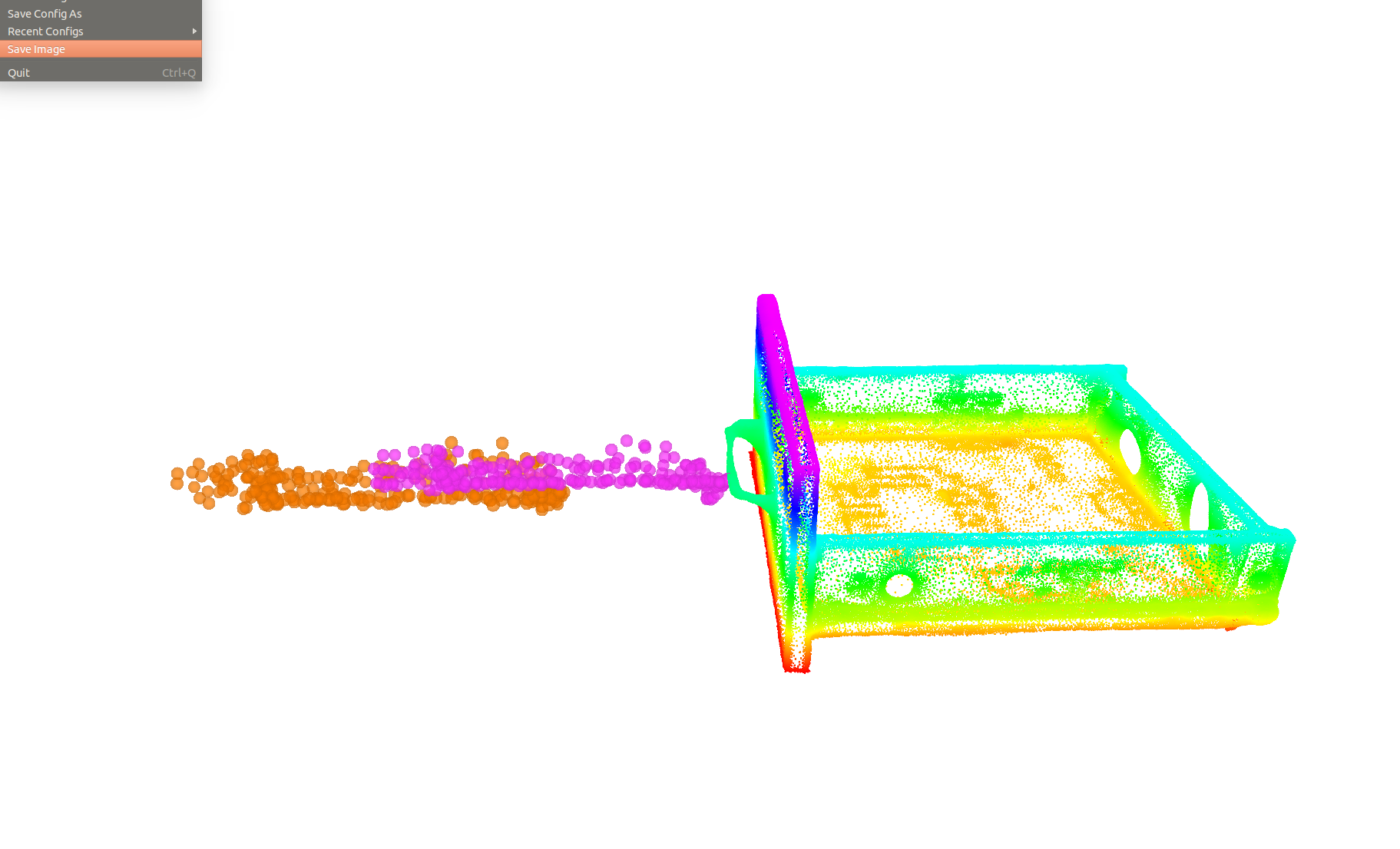} & 
  	\includegraphics[width=0.239\textwidth ,trim={17.0cm 8.5cm 9.5cm 1.5cm},clip]{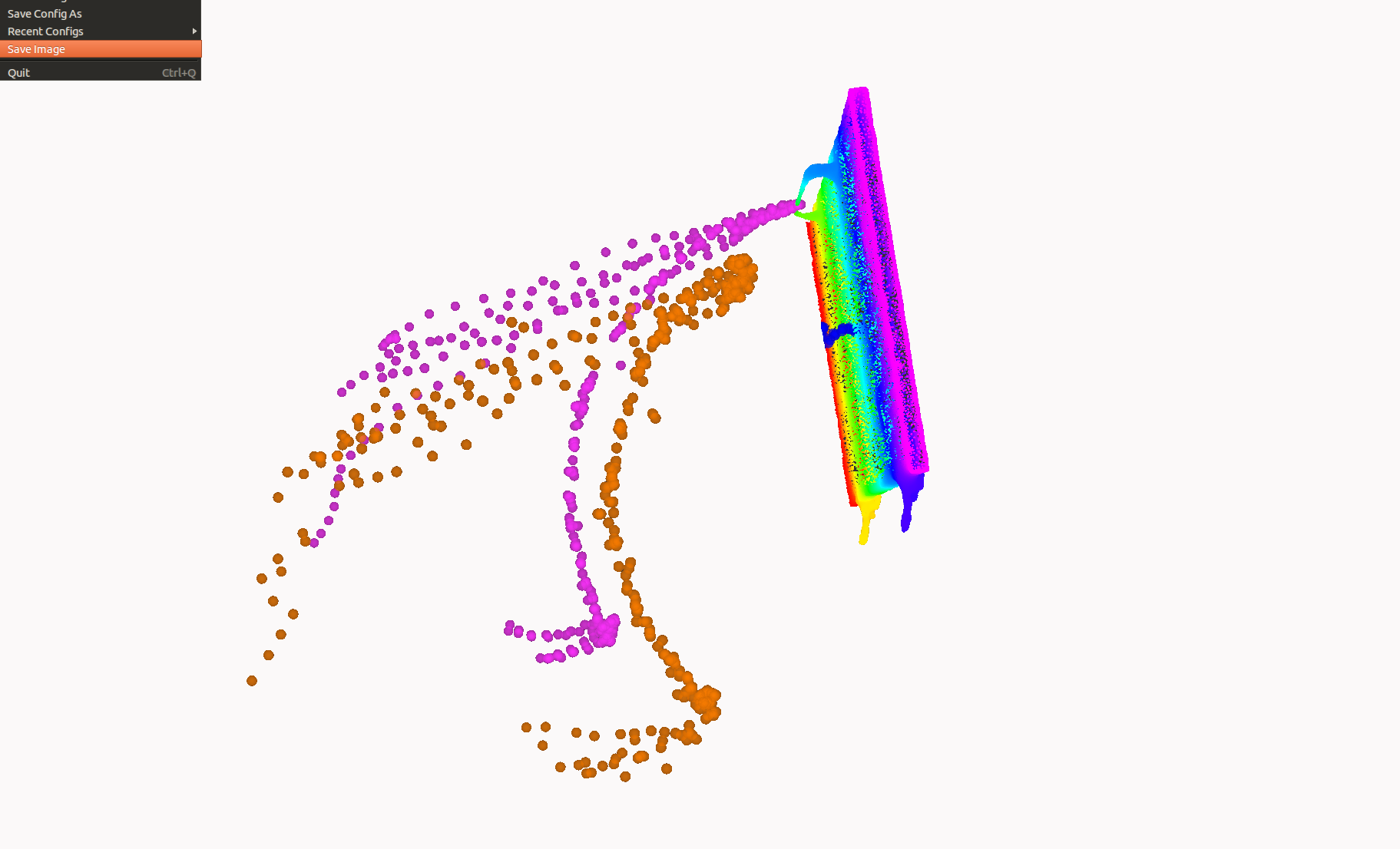}
	\end{tabular}
	\caption{Adaption of the recorded human teacher trajectory to the robot grasping capabilities for grasping the handle of the drawer (left) and the swivel door (right). The gripper poses (magenta dots) are shifted towards the handle of the drawer, respectively the door, leading to a successful robot grasp. By just imitating the human hand motion (orange dots) the grasps would fail.}
  	\label{fig:GraspAdaption}
\end{figure}

First we evaluate adaption of the recorded demonstrations towards the robot capabilities. Specifically we compare the optimization for all aforementioned tasks for two different teacher pose estimation methods. The first relies on detecting markers attached to the teachers hand and torso and was conducted for our previous work~\cite{twelsche17iros}. The second follows the approach presented in this work. In Table~\ref{T:grasp_comparison} we summarize the numerical evaluation for both recording methods. The table shows that the offset between a valid robot grasp and the demonstrated grasp pose is higher for the 3D human pose estimation than for the estimation with markers for all tasks. The highest difference occurs for the room door task, because the hand is occluded in many frames resulting in fewer predictions. Nevertheless our graph-based optimization is still able to shift the human hand trajectory to reproduce the intended grasp, see \figref{fig:GraspAdaption}. This is reflected in higher distances, both Euclidean and angular, between gripper and recorded hand poses after the optimization. Next we compare the standard deviation on the transformations between the object and the gripper, respectively the object and hand in the manipulation segment. These transformations correspond to the robot and human grasps. We see that for both the translational and the rotational part we have comparable values for the two pose estimation methods. This indicates that, although not being as accurate as using markers, we still have a high robustness in the pose estimation, meaning that the error is systematic and the relative measurements are consistent with little deviation. After the optimization we obtain low standard deviations for both the human and the robot grasp, which corresponds, as desired, to a fixed grasp during manipulation. On the one hand the results show that our graph optimization approach is able and stringently necessary to adapt noisy human teacher demonstrations to robot friendly trajectories. On the other hand they also demonstrate that our approach for pose estimation without markers is sufficiently accurate for action learning.

\setlength{\tabcolsep}{5pt}
\begin{table*}
\centering
\begin{tabular}{l | c c | c c | c c | c c |}
&\multicolumn{2}{c|}{Room Door} &\multicolumn{2}{c|}{Swivel Door} &\multicolumn{2}{c|}{Drawer}    &\multicolumn{2}{c|}{Sliding Door}\\
& Before Opt.		& After Opt.			& Before Opt.		& After Opt.			& Before Opt.		& After Opt.			& Before Opt.	& After Opt.\\
\hline
& \multicolumn{8}{c|}{Human Pose Estimation with AR-Marker}\\ 
&\multicolumn{2}{c|}{10 demos, 1529 poses}  &\multicolumn{2}{c|}{4 demos, 419 poses}    &\multicolumn{2}{c|}{6 demos, 656 poses} &\multicolumn{2}{c|}{10 demos, 1482 poses}\\
\hline
 Euclidean distance gripper-grasp	& $\SI{2.82}{cm}$	& $\SI{0.55}{cm}$	& $\SI{2.36}{cm}$	& $\SI{0.49}{cm}$	& $\SI{6.33}{cm}$	& $\SI{0.37}{cm}$	& $\SI{3.23}{cm}$	& $\SI{0.60}{cm}$\\
 Angular distance gripper-grasp		& $18.3\degree$		& $8.0\degree$		& $5.3 \degree$		& $0.7 \degree$		& $5.4\degree$		& $1.6\degree$ 		& $6.5\degree$	& $0.5\degree$\\
 Euclidean distance gripper-hand		& $-$				& $\SI{2.2}{cm}$		& $-$				& $\SI{2.68}{cm}$	& $-$				& $\SI{5.54}{cm}$ 	& $-$	& $\SI{3.13}{cm}$\\
 Angular distance gripper-hand		& $-$				& $13.5 \degree$		& $-$				& $3.0 \degree$		& $-$				& $2.8\degree$ 		& $-$ &$5.8\degree$\\
 Std dev on gripper-object trans.	& $\SI{1.7}{cm}$		& $\SI{0.53}{cm}$	& $\SI{2.35}{cm}$	& $\SI{0.21}{cm}$	& $\SI{2.66}{cm}$		& $\SI{0.18}{cm}$	& $\SI{0.51}{cm}$ 	& $\SI{0.12}{cm}$\\
 Std dev on gripper-object rot.		& $20.5\degree$		& $2.4 \degree$		& $19.3\degree$		& $1.6\degree$		& $0.88\degree$		& $0.21\degree$		& $3.4\degree$ & $0.34\degree$\\
 Std dev on hand-object trans.		& $\SI{1.7}{cm}$		& $\SI{0.5}{cm}$		& $\SI{2.35}{cm}$	& $\SI{0.16}{cm}$	& $\SI{2.66}{cm}$	& $\SI{0.28}{cm}$ 	& $\SI{0.51}{cm}$	& $\SI{0.16}{cm}$\\
 Std dev on hand-object rot.			& $20.5\degree$		& $4.6\degree$		& $19.3\degree$		& $0.9\degree$		& $0.88\degree$		& $0.3\degree$ 		& $3.4\degree$	& $0.6\degree$\\ 
 Map collision free poses			& $\SI{89.2}{\%}$	& $\SI{99.74}{\%}$	& $-$				& $-$				& $-$				& $-$ 				& $-$	& $-$\\
 Kinematically achievable			& $\SI{69.8}{\%}$	& $\SI{96.86}{\%}$	& $\SI{85.9}{\%}$	& $\SI{99.52}{\%}$	& $\SI{87.3}{\%}$	& $\SI{100}{\%}$ 	& $\SI{63.2}{\%}$	& $\SI{99.93}{\%}$\\
 \hline
 & \multicolumn{8}{c|}{3D Human Pose Estimation from RGBD}\\ 
 & \multicolumn{2}{c|}{10 demos, 1215 poses} & \multicolumn{2}{c|}{5 demos, 330 poses} & \multicolumn{2}{c|}{5 demos, 370 poses} &\multicolumn{2}{c|}{5 demos, 451 poses}\\
 \hline
 Euclidean distance gripper-grasp	& $\SI{31.17}{cm}$	& $\SI{0.40}{cm}$	& $\SI{9.77}{cm}$	& $\SI{0.53}{cm}$	& $\SI{16.18}{cm}$	& $\SI{0.32}{cm}$	& $\SI{5.64}{cm}$	& $\SI{0.19}{cm}$\\
 Angular distance gripper-grasp		& $130.27\degree$		& $0.24\degree$		& $102.9\degree$		& $0.4\degree$		& $108.89\degree$		& $0.06\degree$ 		& $149.30\degree$	& $0.07\degree$\\
 Euclidean distance gripper-hand		& $-$				& $\SI{32.78}{cm}$		& $-$				& $\SI{14.19}{cm}$	& $-$				& $\SI{24.18}{cm}$ 	& $-$	& $\SI{19.26}{cm}$\\
 Angular distance gripper-hand		& $-$				& $63.94\degree$		& $-$				& $92.87\degree$		& $-$				& $103.39\degree$ 		& $-$ &$121.69\degree$\\
 Std dev on gripper-object trans.	& $\SI{17.40}{cm}$		& $\SI{0.25}{cm}$	& $\SI{1.75}{cm}$	& $\SI{0.15}{cm}$	& $\SI{1.08}{cm}$		& $\SI{0.12}{cm}$	& $\SI{1.03}{cm}$ 	& $\SI{0.10}{cm}$\\
 Std dev on gripper-object rot.		& $34.31\degree$		& $0.14\degree$		& $23.39\degree$		& $0.76\degree$		& $14.28\degree$		& $0.06\degree$		& $15.86\degree$ & $0.03\degree$\\
 Std dev on hand-object trans.		& $\SI{17.40}{cm}$		& $\SI{8.01}{cm}$		& $\SI{1.75}{cm}$	& $\SI{1.18}{cm}$	& $\SI{1.08}{cm}$	& $\SI{0.83}{cm}$ 	& $\SI{1.03}{cm}$	& $\SI{0.73}{cm}$\\
 Std dev on hand-object rot.			& $34.31\degree$		& $0.50\degree$		& $23.39\degree$		& $0.89\degree$		& $14.28\degree$		& $0.18\degree$ 		& $15.86\degree$	& $0.27\degree$\\ 
 Map collision free poses			& $\SI{89.14}{\%}$	& $\SI{99.51}{\%}$	& $-$				& $-$				& $-$				& $-$ 				& $-$	& $-$\\
 Kinematically achievable			& $\SI{38.10}{\%}$	& $\SI{96.05}{\%}$	& $\SI{80.0}{\%}$	& $\SI{97.27}{\%}$	& $\SI{60.81}{\%}$	& $\SI{98.92}{\%}$ 	& $\SI{63.64}{\%}$	& $\SI{95.12}{\%}$\\
 
\end{tabular}

\caption{Results for the optimization for all four trained tasks. The upper half of the table summarizes the results from experiments conducted in~\cite{twelsche17iros}. There the human pose estimation was obtained using markers. The lower half shows the results of the experiments carried out with the human pose estimation presented in this work. The total number of recorded poses refers to the length after interpolating missing ones. The shown distance between gripper and grasp poses is a mean over the endpoints of the reaching segments of the demonstrations. For the distance between gripper and hand as well as the collisions and the kinematic feasibility all pose tuples are considered. Kinematic feasibility expresses the lookup in the inverse reachability map. For the relation between object and robot gripper respectively human hand a mean over all poses in the manipulation segments is calculated. Since gripper poses are initialized with the measured hand poses no meaningful distance before optimization can be given. For the three furniture operating tasks no collisions with the map are considered.
}
\label{T:grasp_comparison}
\end{table*}
\setlength{\tabcolsep}{6pt}

\setlength{\tabcolsep}{1pt}
\setlength{\tabcolsep}{6pt}

\subsection{Action Imitation by the Robot}
\begin{figure}[t]
	\centering
	\begin{tabular}{cc}
  		\includegraphics[width=0.239\textwidth]{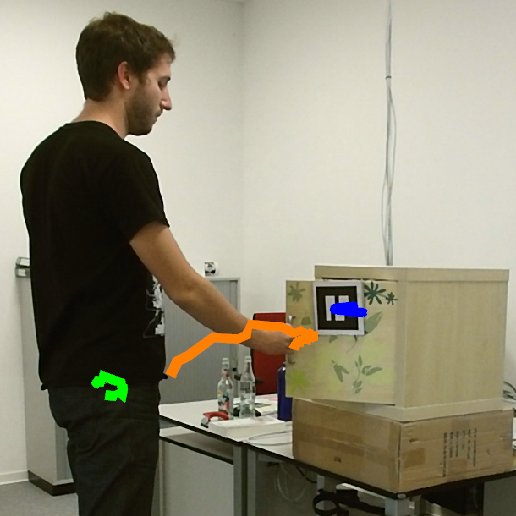} &
  		\includegraphics[width=0.239\textwidth]{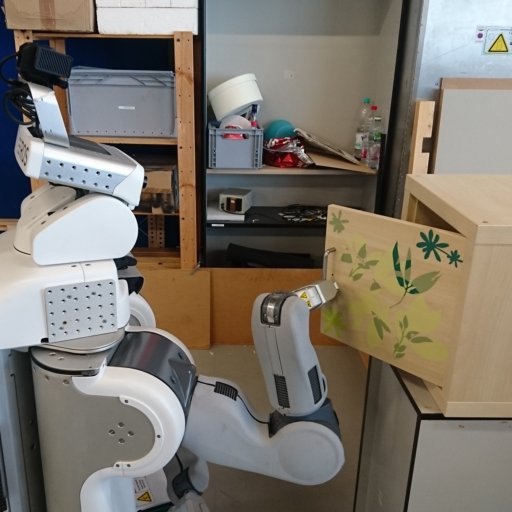} 
	\end{tabular}
	\caption{On the left image the teacher demonstrates the task of opening the swivel door. Superimposed on the image we see the recorded trajectories for hand (orange), torso (green) and manipulated object (blue) which serve as the input for the action learning. The right image shows the robot reproducing the action using a model learned from the teacher demonstration.}
  	\label{fig:robotaction}
\end{figure}

In a follow-up experiment we used the adapted demonstrations from our pose estimation approach shown in Table~\ref{T:grasp_comparison} to learn action models that our PR2 robot can use to imitate the demonstrated actions in real world settings. These time-driven models are learned as in our previous work~\cite{twelsche17iros} using mixtures of Gaussians~\cite{Calinon12Hum}. We learn combined action models for robot gripper and base in Cartesian space. The models are used to generate trajectories for the robot in the frame of the manipulated object.   
With the learned models we reproduced each action five times. For opening the swivel door we had one failure due to localization problems during the grasping. For the drawer and the room door all trials of grasping and manipulating were successful. The sliding door was always grasped successfully but due to the small door knob and the tension resulting from the combined gripper and base motion, the knob was accidentally released during the manipulation process. We ran five successful trials of opening the sliding door by keeping the robot base steady. A visualization of the teaching process and the robot reproducing the action demonstration can be seen in \figref{fig:robotaction}.

\setlength{\tabcolsep}{1pt}
\setlength{\tabcolsep}{6pt}

\section{CONCLUSIONS}
We propose a CNN based system that jointly uses color and depth information in order to predict 3D human pose in real world units. This allows us to exceed the performance of existing methods. Our work introduces two RGBD datasets, which can be used for future approaches. We show, how our approach for 3D human pose estimation is applied in a task learning application that allows non-expert users to teach tasks to service robots. This is demonstrated in real-world experiments that enable our PR2 robot to reproduce human-demonstrated tasks without any markers on the human teacher.

%%%%%%%%%%%%%%%%%%%%%%%%%%%%%%%%%%%% BIBLIOGRAPHY %%%%%%%%%%%%%%%%%%%%%%%%%%%%%%%%%%%%%%%%%%%%

{\small
    \bibliographystyle{ieee}
    \bibliography{humanpose}
}

%%%%%%%%%%%%%%%%%%%%%%%%%%%%%%%%%%%% BIBLIOGRAPHY %%%%%%%%%%%%%%%%%%%%%%%%%%%%%%%%%%%%%%%%%%%%

\addtolength{\textheight}{-12cm}   % This command serves to balance the column lengths on the last page of the document manually. It shortens the textheight of the last page by a suitable amount. This command does not take effect until the next page so it should come on the page before the last. Make sure that you do not shorten the textheight too much.
\end{document}